\pdfoutput=1
\documentclass[letterpaper, 10 pt, conference]{ieeeconf}  

\IEEEoverridecommandlockouts                              

\usepackage[export]{adjustbox}            
\usepackage{multirow}
\usepackage{algorithm}                    
\usepackage[noend]{algpseudocode}         
\usepackage{amsmath}                      
\usepackage{amssymb}                      
\usepackage{cite}                         
\usepackage{graphics}                     
\usepackage[usenames,dvipsnames]{xcolor}  
\usepackage{subfigure}
\usepackage{svg}

\usepackage{url}

\makeatletter
\newcommand\fs@betterruled{%
  \def\@fs@cfont{\bfseries}\let\@fs@capt\floatc@ruled
  \def\@fs@pre{\vspace*{5pt}\hrule height.8pt depth0pt \kern2pt}%
  \def\@fs@post{\kern2pt\hrule\relax}%
  \def\@fs@mid{\kern2pt\hrule\kern2pt}%
  \let\@fs@iftopcapt\iftrue}
\floatstyle{betterruled}
\restylefloat{algorithm}
\def\hlinewd#1{%
\noalign{\ifnum0=`}\fi\hrule \@height #1 %
\futurelet\reserved@a\@xhline}
\makeatother


\newcommand{\algorithmicinput}{\textbf{Input:}}

\newcommand{\INPUT}{\item[\algorithmicinput]}

\newcommand{\CALL}[1]{\textsc{#1}}

\newcommand{\cspace}{\ensuremath{\mathcal{C}_{space}}}
\newcommand{\cspaces}{\ensuremath{\mathcal{C}_{spaces}}}
\newcommand{\cfree}{\ensuremath{\mathcal{C}_{free}}}
\newcommand{\cobst}{\ensuremath{\mathcal{C}_{obst}}}

\newcommand{\blue}[1]{#1}

\title{\LARGE \bf
Scalable Multi-Robot Motion Planning Using\\ Workspace Guidance-Informed Hypergraphs
}

\author{Courtney McBeth$^{1}$, James D. Motes$^{1}$, Isaac Ngui$^{1}$, Marco Morales$^{1}$, and Nancy M. Amato$^{1}$%
\thanks{$^{1}$Courtney McBeth, James D. Motes, Isaac Ngui, Marco Morales, and Nancy M. Amato are with the Parasol Lab, Department of Computer Science,
        University of Illinois Urbana-Champaign, Champaign, IL 61820 USA
        {\tt\small \{cmcbeth2, jmotes2, ingui2, moralesa, namato\}@illinois.edu}}%
\thanks{This work was supported in part by the IBM-Illinois Discovery Accelerator Institute and the Center for Networked Intelligent Components and Environments (C-NICE) at the University of Illinois.
}%
}

\begin{document}

\maketitle
\thispagestyle{empty}
\pagestyle{empty}


\begin{abstract}

In this work, we propose a method for multiple mobile robot motion planning that efficiently plans for robot teams up to \blue{128 robots} (an order of magnitude larger than existing state-of-the-art methods) in congested settings with narrow passages in the environment.
We achieve this improvement in scalability by extending the state-of-the-art Decomposable State Space Hypergraph (DaSH) \blue{multi-robot} planning framework to support \blue{mobile robot motion planning in congested environments. This is a problem that DaSH cannot be directly applied to because it lacks a highly structured, easily discretizable task space and features kinodynamic constraints}.
We accomplish this by exploiting \blue{knowledge about the workspace topology} to limit exploration of the planning space and through modifying DaSH's conflict resolution scheme.
This guidance captures when coordination between robots is necessary, allowing us to decompose the intractably large multi-robot search space while limiting risk of inter-robot conflicts by composing relevant robot groups together while planning.

\end{abstract}


\section{Introduction}

Multi-robot systems are becoming increasingly prevalent in settings like warehouses and factories.
These often constrained environments require careful yet computationally fast motion planning for each robot to accomplish its task without collision.
Some navigation approaches compute individual decoupled paths for each robot and resolve conflicts online~\cite{sl-uppccdpmrs-2002}; however, in congested environments, this introduces the risk of deadlock.
Alternatively, \textit{coupled} offline multi-robot motion planning (MRMP) approaches~\cite{sl-uppccdpmrs-2002} plan for all robots together in the \textit{composite space}, the joint planning space of all robots. While these methods provide the high level of coordination required to solve difficult problems, due to the large size of the search space, they are computationally intractable for large robot teams. \textit{Hybrid} methods~\cite{smsa-romrmpucbs-21, wc-mcmppapb-2011, hpksga-tpqs-2018} seek to combine the benefits of decoupled and coupled planning, providing an increased level of coordination when necessary while also maintaining scalability.
\blue{While existing approaches~\cite{smqma-arc-24, wkc-pppfmrwse-12} reactively provide coordination after conflicts occur during planning, our proposed method aims to improve efficiency by predicting when coordination between robots is necessary.}

\blue{To model when coordination between robots is needed, we build our method off of the Decomposable State Space Hypergraph (DaSH)~\cite{mcbma-hbmrtamp-23} framework.
DaSH presents a general structure for hybrid planning that benefits from a sparse high-level representation of the task space that composes robots into the same planning space when necessary and allows the planner to stay in low-dimensional search spaces when possible.
This representation directs the construction of a motion solution.}
\blue{DaSH proposes an approach to exhaustively constructing the high-level representation that performs well for problems like multi-manipulator rearrangement~\cite{mcbma-hbmrtamp-23}, where the task space has an exploitable structure since actions can be classified as pick, place, handoff, etc., but becomes intractable for problems which lack this highly structured task space. Because of this, we cannot directly apply DaSH to mobile robot motion planning.}

\begin{figure}
    \centering
    \includegraphics[width=0.98\linewidth]{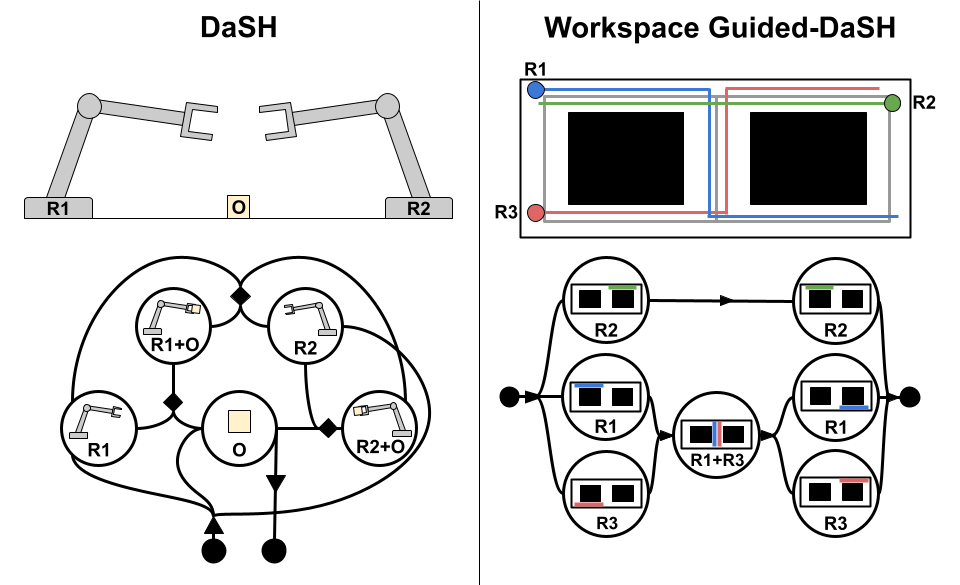}
    \caption{\blue{A comparison of the DaSH framework and Workspace Guided-DaSH showing an example task space hypergraph for each. Hypergraph vertices show which robots and/objects are present in that task space element. (left) DaSH's exhaustive construction of the task space hypergraph for a multi-manipulator rearrangement with two robots and one object. (right) Our method uses workspace guidance to induce structure onto problems that lack this natural structure, which makes construction of a minimal portion of the task space hypergraph possible. Here, robots are composed into the same planning space based on their movement along a topological skeleton (details in Sec.~\ref{sec:method}).}}
    \label{fig:dash-vs-gdash}
\end{figure}

\blue{Here, we propose extending DaSH for mobile robot motion planning in environments with narrow passages, a classical challenge in motion planning exacerbated in the multi-robot setting, where the robots may also have kinodynamic constraints that further complicate planning.}
Our insight is that we can leverage information about the planning problem to induce this structure while building the high-level representation and further exploit this guidance, in conjunction with a modified conflict resolution scheme, to efficiently search the planning space and support kinodynamic constraints.
In mobile robot motion planning, we can effectively use workspace information to provide this guidance~\cite{mmuma-smrmpcetg-23, dsba-drbrrt-16, suda-tgrcdrs-20, rsb-beaeisbmp-2014, pkv-mpdsclp-10}.
As such, we \blue{
extend the DaSH framework to present \textit{Workspace Guided-DaSH}, a novel method for the problem of multiple mobile robot motion planning in environments with narrow passages.}
Our contributions include:
\begin{itemize}
    \item A method that extends the DaSH framework to a multi-robot planning problem without a highly structured task space by leveraging the workspace to guide planning in multiple ways.
    \item Support for planning under kinodynamic constraints via an adapted conflict resolution structure.
    \item An extensive experimental validation showing that our novel approach scales to robot groups up to 128, an order of magnitude larger than previous methods.
\end{itemize}


\section{Preliminaries and Prior Work}
In this section, we describe relevant motion planning preliminaries and discuss prior work in the field of multi-robot motion planning.

\subsection{Motion Planning Preliminaries}
In motion planning, a \textit{configuration} refers to a set of values for a robot's \textit{degrees of freedom}, including their position in the workspace, orientation, and other parameters such as joint angles. The \textit{configuration space} (\cspace{}) consists of all possible configurations and can be partitioned into \cfree{} and \cobst{} made up of valid and invalid configurations, respectively.
Motion planning is the problem of finding a path from a start configuration $q_s$ to a goal configuration $q_g$ through \cfree{}. 
Computing the full \cobst{} is intractable~\cite{ss-otpmpiigtfctporam-83,c-crmp-88}, thus state-of-the-art methods turned to sampling-based motion planning~\cite{kslo-prpp-96,l-rrtntpp-1998}.
Two commonly used methods, Rapidly-exploring Random Trees (RRT)~\cite{l-rrtntpp-1998} and Probabilistic Roadmaps (PRM)~\cite{kslo-prpp-96} build tree-based and graph-based representations of \cfree{} respectively.
These methods have also been extended to problems that require satisfying kinematic and dynamic constraints.
Kinodynamic RRT~\cite{hklr-rkmpwmo-02} extends RRT to work with kinodynamic constraints through extensions of randomly sampled control inputs for random durations until a goal region is reached.

\subsection{Multi-robot Motion Planning}

The MRMP problem is an extension of the motion planning problem that consists of finding valid paths 
$\Pi = \{\pi_{r_1}, ..., \pi_{r_n}\}$ for a set of robots $R = \{r_1, ..., r_n\}$
that also avoid inter-robot collision.
Multi-agent pathfinding (MAPF) is the discrete state space equivalent of the MRMP problem, which consists of finding collision-free paths for a set of robots over a given graph representation (e.g., a grid) rather than through continuous space.
In environments with narrow passages, it is non-trivial to construct a graph representation that captures the connectivity of the planning space, hence our consideration of MRMP \blue{rather than MAPF}.

Some coupled methods, including Composite PRM~\cite{sl-uppccdpmrs-2002} and Composite RRT~\cite{l-rrtntpp-1998}, provide a high level of coordination by planning directly in the composite space, $C_0 \times ... \times C_{n-1}$ where $C_i$ is the configuration space of robot $i$.
Other coupled methods, including MRdRRT~\cite{ssh-faniaehdrfeoirimm-16} and its variants~\cite{dsshb-sammp-17, ssdhb-dsaiammp-20}, construct single-robot roadmaps and then use these to search an implicit composite space roadmap.
\blue{Searching the composite space, even implicitly, becomes intractable with large groups of robots due to the high dimensionality of the space.}

Hybrid methods \blue{alternate between planning in spaces with different compositions of robots.}
CBS-MP~\cite{smsa-romrmpucbs-21} performs a low-level search to find decoupled paths for each robot and resolves conflicts using a high-level search in the composite space.
\blue{Kinodynamic Conflict-Based Search (K-CBS)~\cite{kal-cbsfmrmpwkc-22} extends hybrid MRMP algorithms to accommodate kinodynamic constraints, drawing on the concepts of CBS-MP~\cite{smsa-romrmpucbs-21}.}
Subdimensional Expansion RRT (sRRT)~\cite{wkc-pppfmrwse-12} constructs individual policies by growing RRTs in the \cspace{} of each robot backward from the goal. It then expands a tree forward in the composite space by following the individual policies until conflict occurs. Then, new configurations are sampled for expansion in the composite space of the conflicting robots.
While these methods perform well in open environments, in constrained settings, they expend excess computation on conflict resolution, limiting planning efficiency.
Adaptive Robot Coordination (ARC)~\cite{smqma-arc-24} plans decoupled paths and then resolves conflicts by solving local subproblems in the composite spaces of the affected robots. Though it does focus computation within constrained settings, it does so reactively and spends excess time adjusting the local problem size to find the right resolution. 
\blue{Its kinodynamic variant~\cite{qsmma-karcfmrkp-2025} similarly resolves collisions and ensures kinodynamic feasibility using trajectory optimization.}

\subsection{Guided Motion Planning}
\textit{Guided} motion planning methods use external information to efficiently find paths.
Many \blue{single-robot} approaches use workspace skeletons, embedded graphs in which edges represent free areas of the workspace and vertices are connections between them. Examples include medial axis skeletons~\cite{Blum_1967_6755} for 2D environments and mean curvature skeletons~\cite{t-mcs-12} for 3D. They are generally quick to compute, for example, medial axis skeletons can be computed in $O(n \log n)$ time where $n$ is the number of obstacle edges~\cite{l-matoaps-1982}.
Skeleton-guided planning methods~\cite{dsba-drbrrt-16, suda-tgrcdrs-20} use dynamic sampling regions, bounded local areas of the workspace that advance along skeleton edges, to explore free areas of the environment, including narrow passages, until a path is found.

Prior work extends skeleton guidance to MRMP for scenarios with narrow passages, which require a high level of coordination during planning. Composite Dynamic Region-biased RRT (CDR-RRT)~\cite{mmuma-smrmpcetg-23} is a coupled method that lazily builds and searches over a \textit{composite skeleton}, which is the Cartesian product graph of the skeleton for each robot in the group. MAPF is used to generate a path over the skeleton for each robot, taking into account each edge's capacity as given by the clearance to obstacles in the environment. These paths are combined to form a path through the composite skeleton. Dynamic sampling regions advance over the composite skeleton edges, growing a tree until the goal is reached.
Although CDR-RRT shows a significant improvement in scalability compared to other state-of-the-art methods in environments with narrow passages where coordination between the full robot group is required~\cite{mmuma-smrmpcetg-23}, considering the full composite space of all robots incurs significant computational overhead. We propose a hybrid method extended from CDR-RRT that uses topological guidance to inform when planning in the composite space (of all or a subgroup of robots) is necessary.

\subsection{\blue{Decomposable State Space Hypergraph Framework}}
\label{sec:dash}
\blue{
The Decomposable State Space Hypergraph (DaSH) framework~\cite{mcbma-hbmrtamp-23} for hybrid multi-robot planning leverages a directed hypergraph representation to enable coordination between relevant groups of robots without considering the full composite space unless necessary. A directed hypergraph $\mathcal{H} = (\mathcal{V}, \mathcal{E})$ is a generalization of a graph where hyperarcs $E \in \mathcal{E}$ represent directed connections between sets of vertices.
This representation provides the advantage of sparsity. While an equivalent graph could be constructed, vertices would need to represent a state for every robot in the problem, leading to a combinatorial explosion as the number of robots increases. The hypergraph representation, however, only includes relevant robots in each vertex or hyperarc, allowing groups of robots to be composed and decomposed.
This is especially beneficial for sampling-based motion planning where the planning time is impacted by the dimension of the \cspace{}.}

\blue{
The DaSH framework consists of three components: construction of the \textit{task space hypergraph}, construction of the \textit{motion hypergraph}, and the \textit{query}.
The \textit{task space hypergraph}, $\mathcal{H}_{ts}$, is a high-level representation of the task space that captures when coordination between robots is required. allowing robots to move between planning spaces with different compositions of robots at different stages of planning.
In DaSH, these planning spaces, which are represented by the vertices of the task space hypergraph, are referred to as \textit{task space elements}.
Task space elements are made up of a set of relevant robots and constraints (e.g., start/goal constraints, path constraints).
The DaSH framework has been implemented for problems including multi-manipulator rearrangement~\cite{mcbma-hbmrtamp-23} where the task space is highly structured and easily discretizable with task space elements representing modes such as a robot holding an object. This structure allows the sparse task space hypergraph to be constructed exhaustively. For task spaces without such structure, however, this exhaustive construction becomes intractable.}

\blue{
The task space hypergraph is used to build a low-level \textit{motion hypergraph}, $\mathcal{H}_{m}$, encoding motion paths. Here, vertices represent configurations and hyperarcs represent local paths between them.
The query phase consists of two parts. First, an \textit{optimistic solution} on the motion hypergraph is found which ensures a valid transition history (i.e., guaranteeing that each robot exists in exactly one place at each point along the solution). Second, because hyperarcs in the motion hypergraph give a local plan for the subset of robots in the group being considered and do not make assumptions about the locations of the other robots, a scheduled query must then be used to validate the solution in terms of inter-robot collisions by inducing waiting.}

\blue{
Considering extending DaSH to a problem without a naturally highly structured task space, each of these components poses a design question, which we discuss in Section~\ref{sec:method}.
Additionally, we discuss the modifications to the conflict resolution structure that we make to accommodate kinodynamic planning since we can no longer assume the robots are able to instantaneously change velocity in the scheduled query. These changes allow us to bypass the query stage entirely.}

\begin{algorithm}[t]
\small
  \caption{Workspace Guided-DaSH}
  \label{alg:guided-dash}
  \begin{algorithmic}[1]
    \INPUT Workspace Skeleton $S$, Query $Q$
    \State $\Pi \gets \emptyset$ \Comment{Robot motion paths}
    \State $C \gets \emptyset$ \Comment{MAPF search constraints}
    \While{$\Pi = \emptyset$}
        \State \blue{$\Pi^S \gets \CALL{MAPF}(S, C, Q)$ \Comment{Find paths over the skeleton}\label{line:mapf}}
        \State $\mathcal{H}_{ts} \gets \CALL{ConstructTaskSpaceHypergraph}(\Pi^S)$\label{line:hts}
        \State $\mathcal{X}_{ts} \gets \CALL{OrderByDependency}(\mathcal{H}_{ts})$\label{line:order}
        \State $\mathcal{H}_m \gets \emptyset$ \Comment{Initialize motion hypergraph}
        \For{$x_{ts} \in \mathcal{X}_{ts}$} \Comment{\blue{Iterate over $\mathcal{V}_{ts}$ and $\mathcal{E}_{ts}$}}
            \State  $\pi \gets \CALL{ComputePath}(G, x_{ts})$\label{line:plan}
            \If{$\pi = \emptyset$}\Comment{\blue{Failed to find a path}}
                \State $C \gets C \cup \CALL{GetConstraint}(x_{ts})$\label{line:fail}
                \State \textbf{break} \Comment{Restart heuristic search}
            \EndIf
            \State $\mathcal{X}_{conf} \gets \CALL{FindCollisions}(\mathcal{H}_m, \pi)$\label{line:findcollision}
            \If{$\mathcal{X}_{conf} \neq \emptyset$}\Comment{\blue{Found conflict}}
                \State $C \gets C \cup \CALL{GetConstraints}( \mathcal{X}_{conf} \cup \{x_{ts}\})$\label{line:getconstraints}
                \State \textbf{break} \Comment{Restart heuristic search}
            \EndIf
            \State $\mathcal{H}_m.\CALL{AddPath}(\pi)$\label{line:buildhm}
        \EndFor
        \State $\Pi \gets \CALL{ExtractMotionPaths}(\mathcal{H}_m)$\label{line:extract}
    \EndWhile
  \end{algorithmic}
\end{algorithm}


\section{Workspace Guided-DaSH Method}
\label{sec:method}
In this section, we propose the Workspace Guided-DaSH method, which leverages the workspace skeleton to guide both the construction of $\mathcal{H}_{ts}$ and its translation into $\mathcal{H}_m$.
\blue{Fig.~\ref{fig:dash-vs-gdash} compares the DaSH framework with our method, which we elaborate on below.}

\subsection{\blue{Problem Formulation}}

\blue{We consider the problem of multiple mobile robot motion planning in congested environments with narrow passages. Given a set of $n$ robots, a set of start and goal configurations, $q_s^i~\forall i \in [1, n]$ and $q_g^i~\forall i \in [1, n]$ respectively, we aim to find a path for each robot $\pi^i~\forall i \in [1, n]$ that satisfies the criteria that $\pi^i_0 = q_s^i~\forall i \in [1, n]$ and $\pi^i_T = q_g^i~\forall i \in [1, n]$ where $\pi^i_t$ represents the path configuration at time $t$ and $T$ is the final timestep. Additionally, these paths must be inter-robot collision free. We consider two path configurations to be in collision if the geometry of the robots at these configurations intersect. Our planner is generic and we do not make assumptions about robot geometry or kinematics. We assume a known environment and precompute a workspace skeleton representation.}

\subsection{High Level Task Space Hypergraph Construction}
\blue{
Constructing $\mathcal{H}_{ts}$ requires knowledge about the problem space to make composition decisions about when coordination between robots is required during planning and the relevant path constraints.
}
Considering mobile robots, coordination between robots' motions is required when they are physically near each other. Our insight is that we can leverage the skeleton to approximate when coordination is needed by modeling the movement of the robots through the workspace as movement along the skeleton. Specifically, coordination is needed when robots are traversing the same skeleton edge (either in the same or opposite direction) or traveling through the same skeleton vertex. This formulation imposes a structure on the search space.

\begin{figure}
    \hfill
    \subfigure[Segment of Robot Paths over Skeleton]{
        \includegraphics[width=0.36\textwidth]{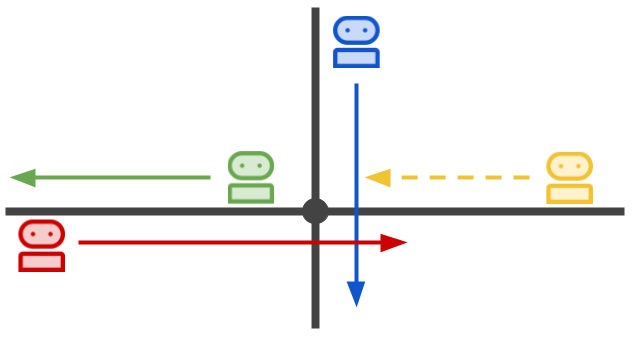}
    }
    \hfill \hfill \\
    
    \subfigure[Generated Portion of Task Space Hypergraph]{
        \includegraphics[width=0.45\textwidth]{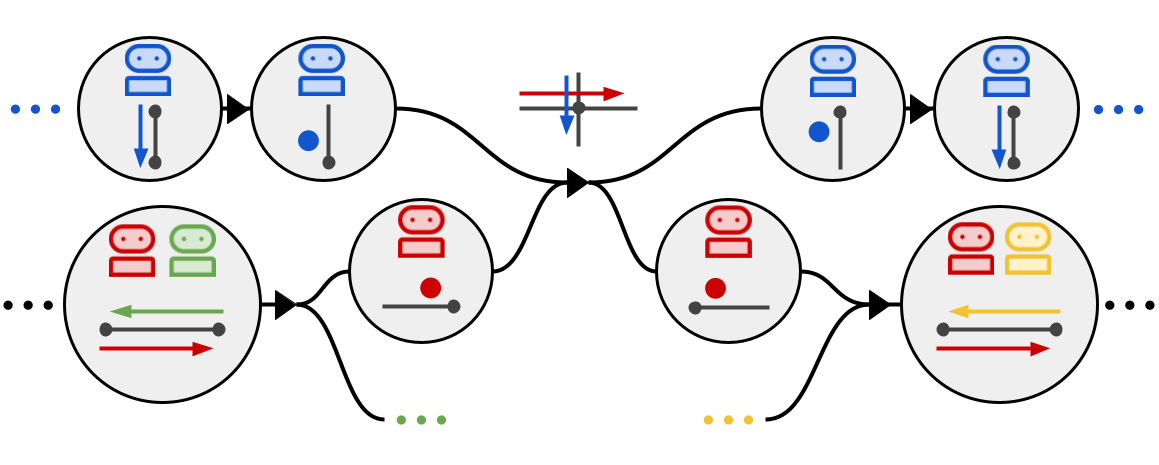}
    }
    \caption{(a) A segment of the robot paths over the skeleton (gray)
    and (b) the corresponding portion of the task space hypergraph. Each task space element \blue{(circle)} shows the corresponding set of robots and the relevant portion of the workspace skeleton. In (b), arrows indicate robots moving along a skeleton edge and dots indicate a decoupled intermediate state.
    The green and red robots are initially moving in opposite directions along the same skeleton edge. Then, red and blue reach the skeleton vertex together and go on to different edges. The yellow robot begins moving in the opposite direction of the red robot when the red robot reaches the next skeleton edge.}
    \label{fig:comp-arcs}
\end{figure}

Workspace Guided-DaSH follows the structure shown in Alg.~\ref{alg:guided-dash}.
Given a workspace skeleton, we first conduct a heuristic search to find a path over the skeleton for each robot using a MAPF algorithm~\cite{ssfs-cbsfomap-15} \blue{(line~\ref{line:mapf})}.
We set the capacity of each skeleton vertex and edge as given by the minimum clearance to an obstacle.
We then combine these paths to form the minimal necessary portion of $\mathcal{H}_{ts}$ \blue{(line~\ref{line:hts})}.
Fig.~\ref{fig:comp-arcs} shows an example of the conversion of the MAPF solution into $\mathcal{H}_{ts}$.
Within task space elements, constraints represent the skeleton segments that each robot must traverse at that stage.
Robots traversing the same skeleton edge are grouped into the same task space element.
As robots move away from each other to different edges, they are decoupled into separate task space elements, boosting planning efficiency by considering smaller \cspaces.

\subsection{Translation to Low-level Motion Hypergraph}
\label{sec:ll_motion}
The motion hypergraph $\mathcal{H}_m$ stores the configurations and motions between them that make up the robots' paths through the task space elements and hyperarcs from $\mathcal{H}_{ts}$.
For each task space element and hyperarc $x_{ts} \in \mathcal{V}_{ts} \cup \mathcal{E}_{ts}$, ordered from start to goal \blue{(line~\ref{line:order})}, a path is computed in the composite space of the robots contained therein \blue{(line~\ref{line:plan})}. These paths become vertices and hyperarcs in $\mathcal{H}_m$.
To ensure that the robots roughly follow their paths over the skeleton and that narrow passages in the environment are efficiently traversed, we leverage skeleton guidance during path construction. To compute paths through task space elements, we use the dynamic sampling region advancement procedure from CDR-RRT~\cite{mmuma-smrmpcetg-23}.

Considering hyperarcs that represent multiple robots moving through a skeleton vertex, we stop the movement of the robots along the incoming skeleton edges a small distance $\delta$ from the end of each edge \blue{to avoid conflicts while moving into the vertex}. We've found that setting $\delta$ to the diameter of the region used for the dynamic sampling region advancement works well. We then grow an RRT~\cite{l-rrtntpp-1998} in the composite space of these robots from the last configuration on the robots' incoming skeleton edges to a sampled configuration at least $\delta$ distance away from the skeleton vertex of the robots' next skeleton edges. Here, we simply grow an RRT rather than using CDR-RRT to avoid forcing the robots to reach the skeleton vertex at the same time, creating congestion.

We consider construction of a path to have failed if the RRT fails to be extended a given number of times during planning \blue{(line~\ref{line:fail})}. In this case, we impose a constraint that the MAPF solution from which we build $\mathcal{H}_{ts}$ cannot contain the same group of robots traversing along either a skeleton edge that failed or the same groups of robots traversing along the incoming or outgoing skeleton edges to a skeleton vertex that failed. We then recompute $\mathcal{H}_{ts}$ \blue{(line~\ref{line:hts})} and try again to construct $\mathcal{H}_m$. 

Since we know from the heuristic search \blue{over the skeleton} which task space elements' motions occur concurrently, we can interleave motion hypergraph construction and conflict detection. By finding conflicts \blue{between different task space elements} as they occur, we minimize wasted computation and can directly extract the motion solution from the motion hypergraph, bypassing DaSH's query stage.
As we construct $\mathcal{H}_m$, after each path is computed, we check for collisions with existing paths occurring simultaneously by considering intermediate configurations along the paths \blue{(line~\ref{line:findcollision})}.
If conflicts are found between robots traversing different portions of the skeleton simultaneously, \blue{similarly to the constraints added when path construction fails, MAPF} constraints are imposed to prevent the relevant robot groups from traversing these segments of the skeleton together \blue{(line~\ref{line:getconstraints})}.

Our modification to the conflict resolution stage also allows us to support kinodynamic planning. The scheduled query proposed in the original DaSH framework resolves conflicts by inducing waiting to prevent robots from colliding with each other. When planning under kinodynamic constraints, we cannot assume the ability of robots to instantaneously change velocity. Additionally, to avoid solving computationally difficult optimal control point-to-point problems, we cannot reuse paths for task space elements occurring after conflicts in the robots' paths. Thus, we must replan affected paths to avoid conflicts.

\blue{After new MAPF constraints are added,} the heuristic search recomputes $\mathcal{H}_{ts}$ \blue{(line~\ref{line:hts})}, and construction of $\mathcal{H}_m$ restarts. This process continues until a conflict-free motion solution is found \blue{(lines~\ref{line:buildhm}-~\ref{line:extract})}.

\subsection{Kinodynamic Planning}
\label{sec:k-wg-dash}

As kinodynamic constraints influence the motion of the robots, it is imperative that the low-level motion hypergraph, $\mathcal{H}_m$, incorporates the kinodynamic constraints during construction.
This is achieved by substituting the underlying RRT with Kinodynamic RRT~\cite{hklr-rkmpwmo-02}.



A control input is generated by sampling the control space uniformly at random, which is then integrated forward during the RRT extension phase for a randomly sampled duration.
In order to utilize the guidance provided by the CDR-RRT, we generate and integrate forward many random control inputs and select the input that moves the system closest to the dynamic sampling region.
The kinodynamic RRT continues to expand until either a maximum number of states have been added, \blue{a failure}, or the goal region has been reached.

\begin{figure}
    \centering
    \subfigure[Warehouse]{
        \includegraphics[width=0.42\textwidth]{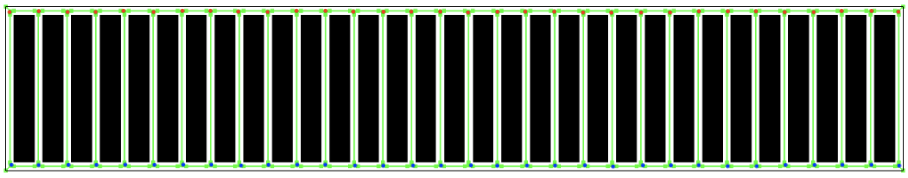}
        \label{fig:warehouse}
    }
    \subfigure[Tunnels Environment]{
        \includegraphics[width=0.38\textwidth]{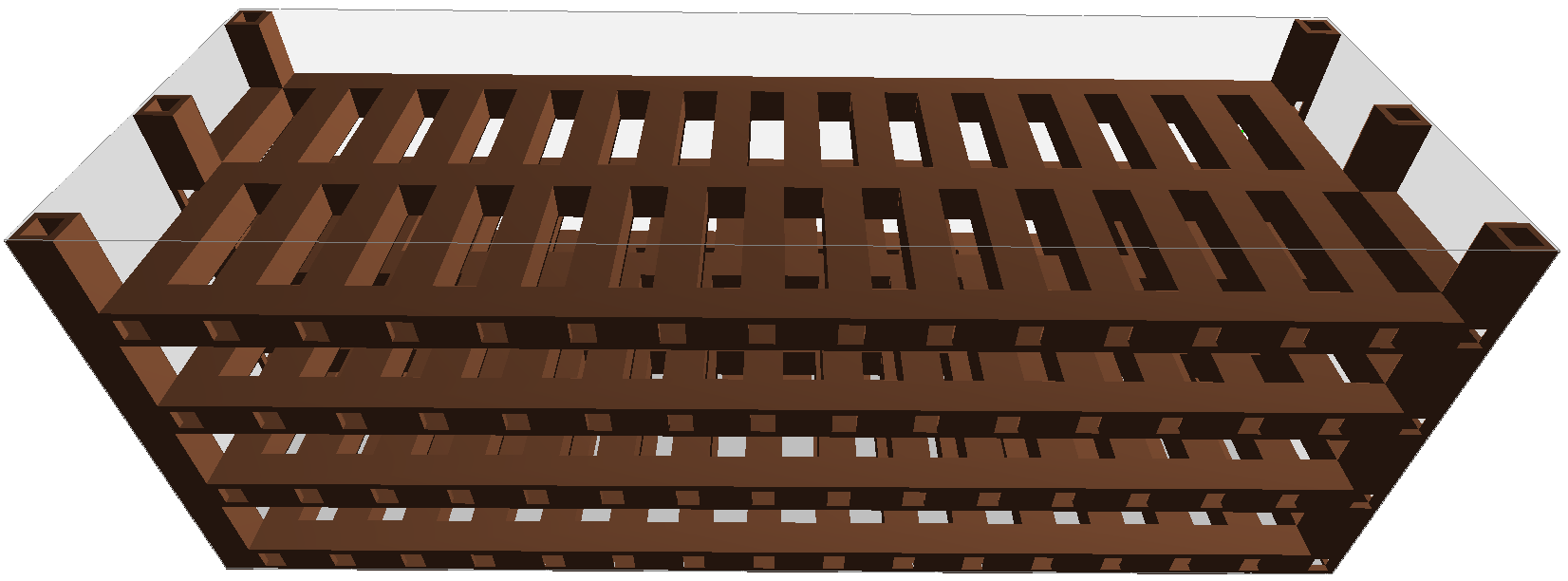}
        \label{fig:tunnels}
    }

    \subfigure[Mining]{
        \includegraphics[width=0.38\textwidth]{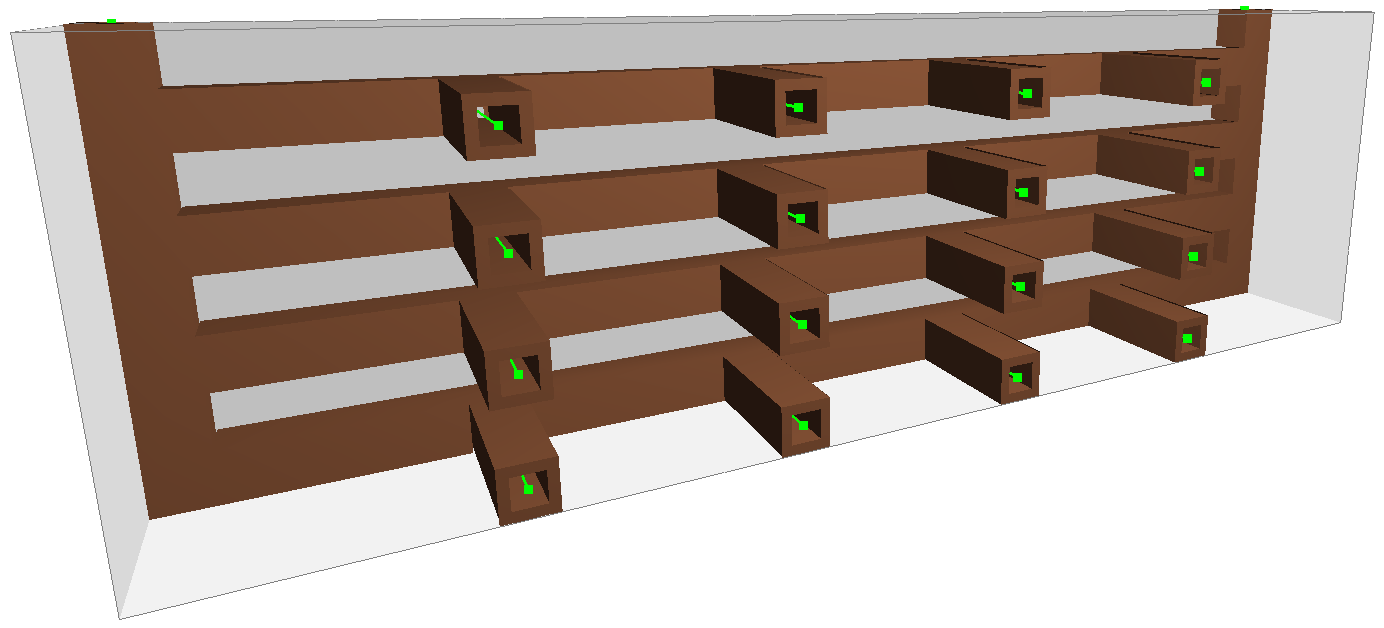}
        \label{fig:mining}
    }
    
    \hfill
    \subfigure[GridMaze]{
        \includegraphics[width=0.15\textwidth]{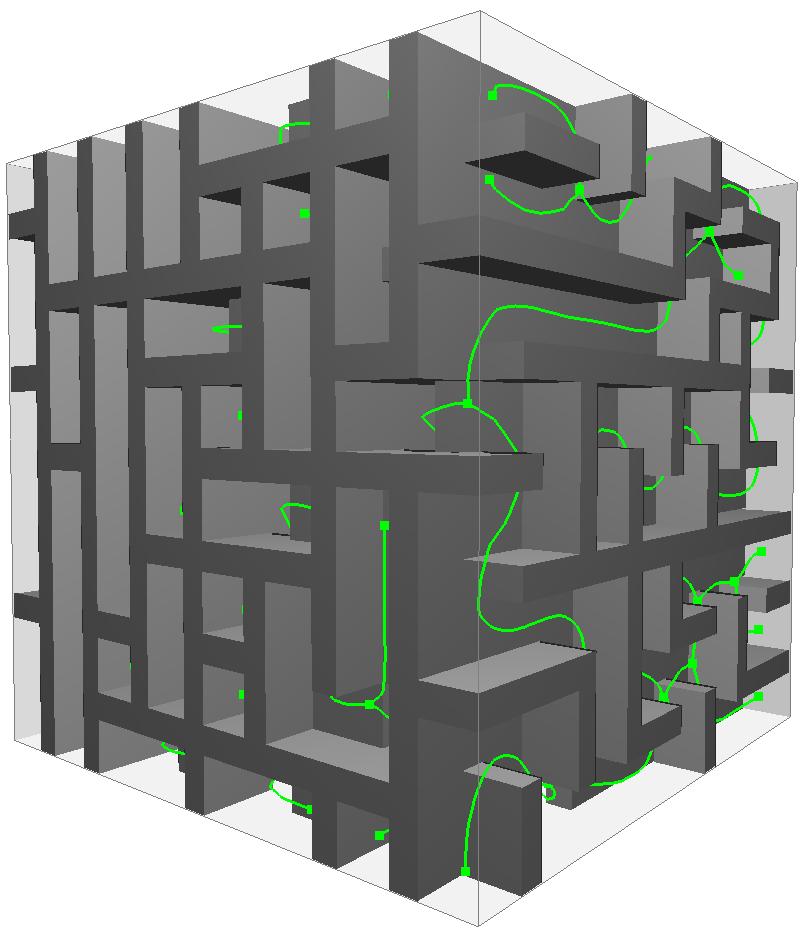}
        \label{fig:gridmaze}
    }
    \hfill
    \subfigure[GridMaze Query]{
        \includegraphics[width=0.15\textwidth]{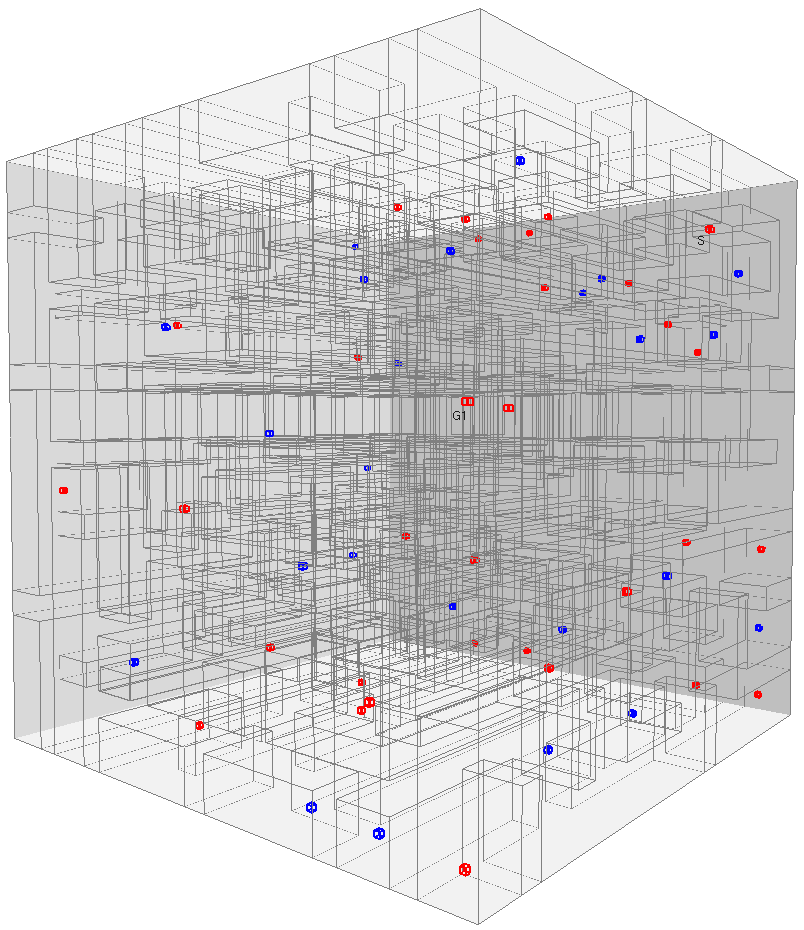}
        \label{fig:gridmaze-query}
    }
    \hfill
    \hfill
    
    \caption{(a) In the Warehouse environment, robots located on the top and bottom of each narrow aisle must swap places with the robot they are vertically aligned with.
    (b) In the Tunnels environment, robots must find and traverse multiple tunnels while moving from start positions in the back tunnel to goal positions in the front tunnel.
    (c) In the mining environment, robots must swap places with another robot located in an adjacent mine shaft, robot start positions are located at the front of the tunnels and at intersections between tunnels.
    (d-e) In the GridMaze environment, robots must move between randomly located starts (red) and goals (blue) within the maze, some of which overlap.}
    \label{fig:exps}
\end{figure}

\section{Validation}

We run scaling MRMP scenarios in environments designed to highlight the strengths and weaknesses of Workspace Guided-DaSH (abbreviated \textit{WG-DaSH} in results) relative to other state-of-the-art approaches.

\subsection{Experimental Setup}
We measure the performance of Workspace Guided-DaSH against CDR-RRT~\cite{mmuma-smrmpcetg-23} as well as a set of other relevant approaches. Composite RRT~\cite{l-rrtntpp-1998} operates in the full composite space, providing high coordination. MRdRRT~\cite{ssh-faniaehdrfeoirimm-16} was originally intended for manipulators, but its implicit search of the composite space makes it pertinent for comparison. 
sRRT~\cite{wkc-pppfmrwse-12} uses subdimensional expansion to limit the size of the search space which is similar in concept to our hypergraph representation.
ARC~\cite{smqma-arc-24} composes robots into higher-dimensional planning spaces based on encountered conflicts, whereas our method uses workspace guidance to preemptively avoid likely conflicts. Thus, comparing our method against ARC effectively measures the impact of this guidance.
Each method was implemented in an open-source library and code will be provided in the final manuscript.
\blue{In all environments, we find plans for holonomic robots considering only geometric feasibility, deferring kinodynamic feasibility to a post-processing trajectory optimization step. In the Warehouse environment, we additionally incorporate kinodynamic constraints into planning.}
We run 15 seeds for each scenario. Each method is given 600 seconds to solve a scenario or is considered a failure. We report planning time as well as path cost given by the makespan.



\begin{table}[t]
\centering
\caption{Warehouse results. Data omitted where all seeds failed}
\label{tab:warehouse}
\begin{tabular}{|l|c|rr|rr|r|}
\hline
\multicolumn{1}{|c|}{\multirow{2}{*}{\textbf{Method}}} & \multirow{2}{*}{\textbf{Robots}} & \multicolumn{2}{c|}{\textbf{Runtime (s)}} & \multicolumn{2}{c|}{\textbf{Cost (s)}} & \multicolumn{1}{c|}{\multirow{2}{*}{\textbf{Success}}} \\ \cline{3-6}
\multicolumn{1}{|c|}{} &  & \multicolumn{1}{c|}{\textbf{Avg}} & \multicolumn{1}{c|}{\textbf{Std}} & \multicolumn{1}{c|}{\textbf{Avg}} & \multicolumn{1}{c|}{\textbf{Std}} & \multicolumn{1}{c|}{} \\ \hline
\textbf{WG-DaSH} & \multirow{6}{*}{2} & \multicolumn{1}{r|}{\textbf{0.7}} & \textbf{0.3} & \multicolumn{1}{r|}{31.6} & 4.4 & 100\% \\ \cline{1-1} \cline{3-7} 
CDR-RRT &  & \multicolumn{1}{r|}{1.5} & 0.8 & \multicolumn{1}{r|}{22.3} & 2.6 & 100\% \\ \cline{1-1} \cline{3-7} 
ARC &  & \multicolumn{1}{r|}{9.6} & 15.3 & \multicolumn{1}{r|}{22.4} & 5.3 & 100\% \\ \cline{1-1} \cline{3-7} 
sRRT &  & \multicolumn{1}{r|}{3.4} & 3.7 & \multicolumn{1}{r|}{29.8} & 9.5 & 100\% \\ \cline{1-1} \cline{3-7}
Comp. RRT &  & \multicolumn{1}{r|}{22.6} & 24.2 & \multicolumn{1}{r|}{25.7} & 12.2 & 100\% \\ \cline{1-1} \cline{3-7}  
MRdRRT &  & \multicolumn{1}{r|}{21.6} & 26.4 & \multicolumn{1}{r|}{82.1} & 35.2 & 53.3\% \\ \hlinewd{0.7pt}
\textbf{WG-DaSH} & \multirow{6}{*}{4} & \multicolumn{1}{r|}{\textbf{1.8}} & \textbf{0.6} & \multicolumn{1}{r|}{32.8} & 3.2 & 100\% \\ \cline{1-1} \cline{3-7} 
CDR-RRT &  & \multicolumn{1}{r|}{2.4} & 1.0 & \multicolumn{1}{r|}{27.6} & 6.7 & 100\% \\ \cline{1-1} \cline{3-7} 
ARC &  & \multicolumn{1}{r|}{16.7} & 18.9 & \multicolumn{1}{r|}{22.8} & 7.9 & 93.3\% \\ \cline{1-1} \cline{3-7} 
sRRT &  & \multicolumn{1}{r|}{14.7} & 9.6 & \multicolumn{1}{r|}{33.2} & 18.9 & 86.7\% \\ \cline{1-1} \cline{3-7} 
Comp. RRT &  & \multicolumn{1}{r|}{274.0} & - & \multicolumn{1}{r|}{19.2} & - & 6.7\% \\ \cline{1-1} \cline{3-7} 
MRdRRT &  & \multicolumn{1}{r|}{295.0} & 233.4 & \multicolumn{1}{r|}{107.1} & 77.4 & 20.0\% \\ \hlinewd{0.7pt}
\textbf{WG-DaSH} & \multirow{4}{*}{8} & \multicolumn{1}{r|}{\textbf{4.6}} & \textbf{3.2} & \multicolumn{1}{r|}{32.5} & 3.0 & 100\% \\ \cline{1-1} \cline{3-7} 
CDR-RRT &  & \multicolumn{1}{r|}{23.0} & 8.2 & \multicolumn{1}{r|}{42.2} & 7.5 & 100\% \\ \cline{1-1} \cline{3-7}
ARC &  & \multicolumn{1}{r|}{37.7} & 49.6 & \multicolumn{1}{r|}{21.0} & 0.9 & 80.0\% \\ \cline{1-1} \cline{3-7}
sRRT &  & \multicolumn{1}{r|}{61.4} & 40.1 & \multicolumn{1}{r|}{20.9} & 0.3 & 13.3\% \\ \hlinewd{0.7pt}
\textbf{WG-DaSH} & \multirow{3}{*}{16} & \multicolumn{1}{r|}{\textbf{9.5}} & \textbf{2.1} & \multicolumn{1}{r|}{31.4} & 3.4 & 100\% \\ \cline{1-1} \cline{3-7} 
CDR-RRT &  & \multicolumn{1}{r|}{288.1} & 100.2 & \multicolumn{1}{r|}{50.1} & 7.1 & 93.3\% \\ \cline{1-1} \cline{3-7}
ARC &  & \multicolumn{1}{r|}{97.1} & 188.5 & \multicolumn{1}{r|}{20.7} & 0.6 & 60.0\% \\ \hlinewd{0.7pt}
\textbf{WG-DaSH} & \multirow{2}{*}{32} & \multicolumn{1}{r|}{\textbf{29.5}} & \textbf{5.8} & \multicolumn{1}{r|}{34.3} & 4.0 & 100\% \\ \cline{1-1} \cline{3-7}
ARC &  & \multicolumn{1}{r|}{72.4} & 7.1 & \multicolumn{1}{r|}{20.9} & 0.7 & 26.6\% \\ \hlinewd{0.7pt}
\textbf{WG-DaSH} & 64 & \multicolumn{1}{r|}{\textbf{54.1}} & \textbf{11.4} & \multicolumn{1}{r|}{35.6} & 2.7 & 100\% \\ \hlinewd{0.7pt}
\textbf{WG-DaSH} & 128 & \multicolumn{1}{r|}{\textbf{250.7}} & \textbf{29.4} & \multicolumn{1}{r|}{36.6} & 5.0 & 60.0\% \\ \hline
\end{tabular}
\end{table}

\subsection{Environments}
Here, we describe our experimental environments and explain which aspects of our approach they highlight.

\subsubsection{Warehouse (Fig.~\ref{fig:warehouse})}
\label{sec:warehouse-setup}
The Warehouse environment features narrow aisles through which the robots must swap places. In the holonomic scenario, the width of the narrow passages is 2.5 times the diameter of the disk robots we consider. We additionally run a kinodynamic variant with doubled aisle widths. This scenario measures each approach's ability to efficiently handle narrow passages within the environment. We highlight Workspace Guided-DaSH's ability to decompose the problem using topological guidance.

\subsubsection{Tunnels (Fig.~\ref{fig:tunnels})}
The Tunnels environment again features narrow passages. Because of the 3D workspace, the planning space becomes very large. This scenario is intended to measure each algorithm's performance with respect to the geometric complexity, rather than the congestion, of the environment. We consider L-shaped robots with the width of their longest dimension slightly less than half the width of the narrow passages. Each robot must find and traverse multiple narrow passages to reach its goal, but its shortest path does not conflict with any other robot's.


\begin{table}[]
\centering
\caption{Tunnels results. Data omitted where all seeds failed}
\label{tab:tunnels}
\begin{tabular}{|l|c|rr|rr|r|}
\hline
\multicolumn{1}{|c|}{\multirow{2}{*}{\textbf{Method}}} & \multirow{2}{*}{\textbf{Robots}} & \multicolumn{2}{c|}{\textbf{Runtime (s)}} & \multicolumn{2}{c|}{\textbf{Cost (s)}} & \multicolumn{1}{c|}{\multirow{2}{*}{\textbf{Success}}} \\ \cline{3-6}
\multicolumn{1}{|c|}{} &  & \multicolumn{1}{c|}{\textbf{Avg}} & \multicolumn{1}{c|}{\textbf{Std}} & \multicolumn{1}{c|}{\textbf{Avg}} & \multicolumn{1}{c|}{\textbf{Std}} & \multicolumn{1}{c|}{} \\ \hline
\textbf{WG-DaSH} & \multirow{3}{*}{2} & \multicolumn{1}{r|}{\textbf{4.2}} & \textbf{0.9} & \multicolumn{1}{r|}{166.0} & 5.1 & 100\% \\ \cline{1-1} \cline{3-7} 
CDR-RRT &  & \multicolumn{1}{r|}{6.2} & 1.1 & \multicolumn{1}{r|}{140.6} & 11.2 & 100\% \\ \cline{1-1} \cline{3-7} 
sRRT &  & \multicolumn{1}{r|}{129.6} & 175.0 & \multicolumn{1}{r|}{175.3} & 26.1 & 40.0\% \\ \hlinewd{0.7pt}
\textbf{WG-DaSH} & \multirow{3}{*}{4} & \multicolumn{1}{r|}{\textbf{8.2}} & \textbf{1.0} & \multicolumn{1}{r|}{158.3} & 9.0 & 100\% \\ \cline{1-1} \cline{3-7} 
CDR-RRT &  & \multicolumn{1}{r|}{17.3} & 3.3 & \multicolumn{1}{r|}{152.9} & 16.4 & 100\% \\ \cline{1-1} \cline{3-7} 
sRRT &  & \multicolumn{1}{r|}{327.0} & - & \multicolumn{1}{r|}{161.7} & - & 6.7\% \\ \hlinewd{0.7pt}
\textbf{WG-DaSH} & \multirow{2}{*}{8} & \multicolumn{1}{r|}{\textbf{14.1}} & \textbf{2.4} & \multicolumn{1}{r|}{159.8} & 8.4 & 100\% \\ \cline{1-1} \cline{3-7} 
CDR-RRT &  & \multicolumn{1}{r|}{36.0} & 6.8 & \multicolumn{1}{r|}{156.1} & 9.0 & 100\% \\ \hlinewd{0.7pt}
\textbf{WG-DaSH} & \multirow{2}{*}{16} & \multicolumn{1}{r|}{\textbf{22.5}} & \textbf{1.9} & \multicolumn{1}{r|}{164.8} & 9.4 & 100\% \\ \cline{1-1} \cline{3-7} 
CDR-RRT &  & \multicolumn{1}{r|}{73.1} & 8.5 & \multicolumn{1}{r|}{163.9} & 16.7 & 100\% \\ \hlinewd{0.7pt}
\textbf{WG-DaSH} & \multirow{2}{*}{32} & \multicolumn{1}{r|}{\textbf{54.0}} & \textbf{6.2} & \multicolumn{1}{r|}{163.0} & 8.0 & 100\% \\ \cline{1-1} \cline{3-7} 
CDR-RRT &  & \multicolumn{1}{r|}{252.6} & 10.0 & \multicolumn{1}{r|}{159.8} & 4.7 & 93.3\% \\ \hlinewd{0.7pt}
\textbf{WG-DaSH} & 64 & \multicolumn{1}{r|}{\textbf{132.4}} & \textbf{7.0} & \multicolumn{1}{r|}{163.1} & 6.0 & 100\% \\ \hline
\end{tabular}
\end{table}

\subsubsection{Mining (Fig.~\ref{fig:mining})}
The Mining environment is a 3D environment meant to mimic a set of mine shafts. Again, it features narrow passages; however, here robots must move in close proximity to each other. We use the same robots as the Tunnels environment. We show that Workspace Guided-DaSH is capable of efficiently finding paths for large robot teams in problems representative of real-world scenarios.

\subsubsection{GridMaze (Fig.~\ref{fig:gridmaze})}
The GridMaze environment is a 3D environment featuring several intersecting narrow tunnels intended to elicit high congestion for large groups of robots in addition to geometric complexity. We use a rectangular prism robot whose smallest dimension is slightly less than half the width of the narrow passages. We demonstrate Workspace Guided-DaSH's ability to efficiently find paths for large groups of robots in these congested settings.


\begin{table}[]
\centering
\caption{Mining results. Data omitted where all seeds failed}
\label{tab:mining}
\begin{tabular}{|l|c|rr|rr|r|}
\hline
\multicolumn{1}{|c|}{\multirow{2}{*}{\textbf{Method}}} & \multirow{2}{*}{\textbf{Robots}} & \multicolumn{2}{c|}{\textbf{Runtime (s)}} & \multicolumn{2}{c|}{\textbf{Cost (s)}} & \multicolumn{1}{c|}{\multirow{2}{*}{\textbf{Success}}} \\ \cline{3-6}
\multicolumn{1}{|c|}{} &  & \multicolumn{1}{c|}{\textbf{Avg}} & \multicolumn{1}{c|}{\textbf{Std}} & \multicolumn{1}{c|}{\textbf{Avg}} & \multicolumn{1}{c|}{\textbf{Std}} & \multicolumn{1}{c|}{} \\ \hline
\textbf{WG-DaSH} & \multirow{3}{*}{2} & \multicolumn{1}{r|}{\textbf{1.3}} & \textbf{0.3} & \multicolumn{1}{r|}{125.6} & 6.5 & 100\% \\ \cline{1-1} \cline{3-7} 
CDR-RRT &  & \multicolumn{1}{r|}{1.9} & 0.4 & \multicolumn{1}{r|}{113.0} & 11.7 & 100\% \\ \cline{1-1} \cline{3-7} 
sRRT &  & \multicolumn{1}{r|}{146.8} & 161.7 & \multicolumn{1}{r|}{131.5} & 17.5 & 33.3\% \\ \hlinewd{0.7pt}
\textbf{WG-DaSH} & \multirow{2}{*}{4} & \multicolumn{1}{r|}{\textbf{2.4}} & \textbf{0.5} & \multicolumn{1}{r|}{122.6} & 10.6 & 100\% \\ \cline{1-1} \cline{3-7} 
CDR-RRT &  & \multicolumn{1}{r|}{16.0} & 6.1 & \multicolumn{1}{r|}{151.2} & 19.5 & 100\% \\ \hlinewd{0.7pt}
\textbf{WG-DaSH} & \multirow{2}{*}{8} & \multicolumn{1}{r|}{\textbf{5.0}} & \textbf{0.8} & \multicolumn{1}{r|}{132.2} & 6.7 & 100\% \\ \cline{1-1} \cline{3-7} 
CDR-RRT &  & \multicolumn{1}{r|}{45.7} & 14.8 & \multicolumn{1}{r|}{160.6} & 12.9 & 100\% \\ \hlinewd{0.7pt}
\textbf{WG-DaSH} & \multirow{2}{*}{16} & \multicolumn{1}{r|}{\textbf{11.0}} & \textbf{1.0} & \multicolumn{1}{r|}{130.3} & 5.8 & 100\% \\ \cline{1-1} \cline{3-7} 
CDR-RRT &  & \multicolumn{1}{r|}{118.9} & 44.3 & \multicolumn{1}{r|}{179.0} & 17.5 & 100\% \\ \hlinewd{0.7pt}
\textbf{WG-DaSH} & 32 & \multicolumn{1}{r|}{\textbf{28.1}} & \textbf{3.5} & \multicolumn{1}{r|}{127.6} & 6.7 & 100\% \\ \hlinewd{0.7pt}
\textbf{WG-DaSH} & 64 & \multicolumn{1}{r|}{\textbf{80.2}} & \textbf{6.9} & \multicolumn{1}{r|}{124.5} & 4.5 & 86.7\% \\ \hline
\end{tabular}
\end{table}

\subsection{Experimental Results}
In the Warehouse scenario (Tab.~\ref{tab:warehouse}), all methods except MRdRRT achieve a 100\% success rate in the 2-robot scenario. However, the methods that do not leverage guidance begin to struggle with the 4-robot scenario and multiple cannot solve the 8-robot scenario.
Of those, only ARC is able to solve up to the 32-robot scenario; however, with a success rate of only 26.6\%.
CDR-RRT and sRRT are able to solve up to the 8-robot scenario where the exponential size of the composite space leads to a significant decrease in performance. 
Workspace Guided-DaSH, like CDR-RRT, sees the benefit of using the skeleton representation to guide planning through narrow passages and the coordination provided by using the composite space of multiple robots.
Additionally, by using the workspace to inform the planner when to move between different levels of composition, Workspace Guided-DaSH is able to focus planning within smaller spaces than CDR-RRT, making it able to plan for significantly larger robot teams, up to 128 robots, within the 600-second time limit. \blue{In this largest 128 robot scenario, the runtime of our MAPF heuristic was $16.7 \pm 6.4$ seconds, demonstrating that it maintains scalability for large robot groups as well.}

In the Tunnels environment (Tab.~\ref{tab:tunnels}), the size of the search space is increased significantly relative to the Warehouse environment since the robot now has six degrees of freedom. Composite RRT, MRdRRT, and ARC fail to solve the 2-robot scenario due to the complexity of the environment. sRRT performs better because of its ability to decompose the search space, but struggles with narrow passages and only one seed was able to solve the 4-robot scenario. Here, the narrow passage problem is exacerbated since the query requires the planner to find and traverse multiple narrow passages. By leveraging skeleton guidance, CDR-RRT and Workspace Guided-DaSH are able to find solutions for larger groups of robots. Searching the composite space hinders CDR-RRT's performance, leaving it unable to solve the 64-robot scenario within the time limit. Workspace Guided-DaSH is able to efficiently find solutions for up to 64 robots within the time limit due to its ability to decompose the planning space.


\begin{table}[]
\centering
\caption{GridMaze results. Data omitted where all seeds failed}
\label{tab:gridmaze}
\begin{tabular}{|l|c|rr|rr|r|}
\hline
\multicolumn{1}{|c|}{\multirow{2}{*}{\textbf{Method}}} & \multirow{2}{*}{\textbf{Robots}} & \multicolumn{2}{c|}{\textbf{Runtime (s)}} & \multicolumn{2}{c|}{\textbf{Cost (s)}} & \multicolumn{1}{c|}{\multirow{2}{*}{\textbf{Success}}} \\ \cline{3-6}
\multicolumn{1}{|c|}{} &  & \multicolumn{1}{c|}{\textbf{Avg}} & \multicolumn{1}{c|}{\textbf{Std}} & \multicolumn{1}{c|}{\textbf{Avg}} & \multicolumn{1}{c|}{\textbf{Std}} & \multicolumn{1}{c|}{} \\ \hline
\textbf{WG-DaSH} & \multirow{5}{*}{2} & \multicolumn{1}{r|}{\textbf{0.3}} & \textbf{0.04} & \multicolumn{1}{r|}{44.0} & 5.5 & 100\% \\ \cline{1-1} \cline{3-7} 
CDR-RRT &  & \multicolumn{1}{r|}{1.0} & 0.3 & \multicolumn{1}{r|}{63.9} & 8.3 & 100\% \\ \cline{1-1} \cline{3-7} 
ARC &  & \multicolumn{1}{r|}{146.4} & 78.2 & \multicolumn{1}{r|}{25.7} & 2.0 & 100\% \\ \cline{1-1} \cline{3-7} 
sRRT &  & \multicolumn{1}{r|}{4.7} & 4.8 & \multicolumn{1}{r|}{30.7} & 3.8 & 100\% \\ \cline{1-1} \cline{3-7}
Comp. RRT &  & \multicolumn{1}{r|}{425.0} & 136.0 & \multicolumn{1}{r|}{59.4} & 11.5 & 20.0\% \\ \hlinewd{0.7pt}
\textbf{WG-DaSH} & \multirow{4}{*}{4} & \multicolumn{1}{r|}{\textbf{1.0}} & \textbf{0.5} & \multicolumn{1}{r|}{44.2} & 3.8 & 100\% \\ \cline{1-1} \cline{3-7} 
CDR-RRT &  & \multicolumn{1}{r|}{46.0} & 17.7 & \multicolumn{1}{r|}{120.6} & 15.9 & 100\% \\ \cline{1-1} \cline{3-7}
ARC &  & \multicolumn{1}{r|}{215.2} & 137.6 & \multicolumn{1}{r|}{26.3} & 2.1 & 93.3\% \\ \cline{1-1} \cline{3-7} 
sRRT &  & \multicolumn{1}{r|}{19.2} & 6.0 & \multicolumn{1}{r|}{30.9} & 4.1 & 100\% \\ \hlinewd{0.7pt}
\textbf{WG-DaSH} & \multirow{3}{*}{8} & \multicolumn{1}{r|}{\textbf{3.2}} & \textbf{1.5} & \multicolumn{1}{r|}{42.3} & 3.3 & 100\% \\ \cline{1-1} \cline{3-7}
ARC &  & \multicolumn{1}{r|}{361.4} & 144.2 & \multicolumn{1}{r|}{26.4} & 2.6 & 93.3\% \\ \cline{1-1} \cline{3-7} 
sRRT &  & \multicolumn{1}{r|}{25.5} & 10.6 & \multicolumn{1}{r|}{32.8} & 9.7 & 100\% \\ \hlinewd{0.7pt}
\textbf{WG-DaSH} & \multirow{3}{*}{16} & \multicolumn{1}{r|}{\textbf{3.8}} & \textbf{1.6} & \multicolumn{1}{r|}{44.2} & 4.1 & 100\% \\ \cline{1-1} \cline{3-7}
ARC &  & \multicolumn{1}{r|}{437.0} & 55.9 & \multicolumn{1}{r|}{26.0} & 1.9 & 40.0\% \\ \cline{1-1} \cline{3-7} 
sRRT &  & \multicolumn{1}{r|}{38.1} & 10.8 & \multicolumn{1}{r|}{29.3} & 2.6 & 100\% \\ \hlinewd{0.7pt}
\textbf{WG-DaSH} & 32 & \multicolumn{1}{r|}{\textbf{25.5}} & \textbf{7.9} & \multicolumn{1}{r|}{36.3} & 16.4 & 93.3\% \\ \hline
\end{tabular}
\end{table}

In the Mining environment (Tab.~\ref{tab:mining}), Composite RRT and MRdRRT fail to solve the 2-robot scenario due to the large size of the search space and inter-robot conflicts.
Similar to the Tunnels scenario, ARC's performance is impacted by the environment complexity since it does not use topological guidance and it fails to solve the 2-robot scenario.
sRRT struggles to solve the 2-robot scenario and cannot solve the 4-robot scenario. By leveraging topological guidance to limit exploration of the composite space, CDR-RRT is able to solve up to the 16-robot scenario. However, it is unable to solve the 32-robot scenario. Workspace Guided-DaSH, by considering the composite space only when necessary, is able to efficiently solve up to the 64-robot scenario.

In the 3D GridMaze scenario (Tab.~\ref{tab:gridmaze}), as a result of the large size of the planning space as well as the difficulty posed by the long curved narrow passages, Composite RRT struggles with the 2-robot scenario and fails to solve the 4-robot scenario. MRdRRT fails to solve the 2-robot scenario. While CDR-RRT is able to solve the 4-robot scenario, the large size of the composite space prevents it from solving the 8-robot scenario within the time limit.
sRRT and ARC benefit from searching a lower-dimensional space when robots are not in conflict, but fail to solve the 32-robot scenario where conflicts become prevalent due to the high level of congestion of the environment. 
Similar to sRRT, Workspace Guided-DaSH searches lower-dimensional spaces, however, our method also uses the workspace skeleton for conflict resolution. When a conflict is too difficult to be resolved quickly via random sampling, we impose constraints on the robots' paths over the workspace skeleton which allows us to find conflict-free paths more efficiently. Thus, Workspace Guided-DaSH is able to find solutions up to 32 robots. Although this replanning mechanism sometimes results in slightly higher solution costs relative to other methods, a post-processing trajectory optimization step is generally used in practice to smooth paths.

\subsection{Kinodynamic Experiments}
To evaluate Workspace Guided-DaSH with kinodynamic constraints, we use Kinodynamic Workspace Guided-DaSH (K-WG-DaSH) incorporating the modifications in Sec.~\ref{sec:k-wg-dash}.
\blue{We compare our approach to Kinodynamic-CBS~\cite{kal-cbsfmrmpwkc-22} (K-CBS) and Kinodynamic-ARC~\cite{qsmma-karcfmrkp-2025} (K-ARC).}
K-CBS uses the same Kinodynamic RRT we employ in K-WG-DaSH, however, they utilize a merge bound~\cite{ssfs-macbsfomapp-12} creating composite robots when more coordination is needed.
\blue{K-ARC uses the same reactive coordination approach employed by ARC, but additionally uses trajectory optimization to ensure kinodynamic feasibility.}
These allow us to evaluate the benefits of the guidance provided by K-WG-DaSH.
Each method was given $45$ minutes to find a solution to planning problems in the warehouse environment, described in Section~\ref{sec:warehouse-setup}, using robots with $2$nd-order car dynamics:
\begin{equation}
    \dot{x} = v \cos{\theta}, \
    \dot{y} = v \sin{\theta}, \
    \dot{\theta} = \frac{v}{l} \tan{\phi}, \
    \dot{v} = a, \
    \dot{\phi} = \gamma \
\end{equation}


\blue{Results for the kinodynamic warehouse environment are shown in Tab.~\ref{tab:kino-warehouse}.}
\blue{K-ARC was unable to solve the 2-robot scenario within the time limit and is omitted from Tab.~\ref{tab:kino-warehouse}. K-ARC struggles in this scenario because of both the large size of the environment and the constrained space in which both robots must move, which make the trajectory optimization problem difficult to solve.}
K-WG-DaSH finds solutions for problems with up to 6 robots consistently while K-CBS begins to struggle at 4 robots.
In the 2-robot scenario, K-WG-DaSH maintains a high success rate with a lower planning time at the expense of a higher path cost as opposed to K-CBS which is inconsistent at finding solutions with a higher planning time but with a lower path cost.
We attribute the scalability to the topological guidance provided by K-WG-DaSH.



\begin{table}[]
\centering
\caption{Kinodynamic Warehouse results. Data omitted where all seeds failed}
\label{tab:kino-warehouse}
\begin{tabular}{|l|c|rr|rr|r|}
\hline
\multicolumn{1}{|c|}{\multirow{2}{*}{\textbf{Method}}} & \multirow{2}{*}{\textbf{Robots}} & \multicolumn{2}{c|}{\textbf{Runtime (s)}} & \multicolumn{2}{c|}{\textbf{Cost (s)}} & \multicolumn{1}{c|}{\multirow{2}{*}{\textbf{Success}}} \\ \cline{3-6}
\multicolumn{1}{|c|}{} &  & \multicolumn{1}{c|}{\textbf{Avg}} & \multicolumn{1}{c|}{\textbf{Std}} & \multicolumn{1}{c|}{\textbf{Avg}} & \multicolumn{1}{c|}{\textbf{Std}} & \multicolumn{1}{c|}{} \\ \hline
\textbf{K-WG-DaSH} & \multirow{2}{*}{2} & \multicolumn{1}{r|}{\textbf{201.2}} & \textbf{119.1} & \multicolumn{1}{r|}{46.7} & 9.9 & \textbf{93.3\%} \\ \cline{1-1} \cline{3-7} 
K-CBS &  & \multicolumn{1}{r|}{390.1} & 365.9 & \multicolumn{1}{r|}{34.4} & 2.3 & 46.7\% \\ \hlinewd{0.7pt}
\textbf{K-WG-DaSH} & \multirow{2}{*}{4} & \multicolumn{1}{r|}{580.2} & \textbf{376.2} & \multicolumn{1}{r|}{41.9} & 7.3 & \textbf{93.3\%} \\ \cline{1-1} \cline{3-7} 
K-CBS &  & \multicolumn{1}{r|}{\textbf{445.0}} & 473.3 & \multicolumn{1}{r|}{35.1} & 2.7 & 20.0\% \\ \hlinewd{0.7pt}
\textbf{K-WG-DaSH} & \multirow{2}{*}{6} & \multicolumn{1}{r|}{1261.6} & \textbf{524.5} & \multicolumn{1}{r|}{32.1} & 14.8 & \textbf{86.7\%} \\ \cline{1-1} \cline{3-7}
K-CBS &  & \multicolumn{1}{r|}{\textbf{56.8}} & - & \multicolumn{1}{r|}{33.5} & - & 6.7\% \\ \hlinewd{0.7pt}
\textbf{K-WG-DaSH} & \multirow{1}{*}{8} & \multicolumn{1}{r|}{\textbf{2147.2}} & \textbf{313.1} & \multicolumn{1}{r|}{19.1} & 16.0 & \textbf{33.3\%} \\ \hlinewd{0.7pt}
\textbf{K-WG-DaSH} & 10 & \multicolumn{1}{r|}{\textbf{2544.7}} & - & \multicolumn{1}{r|}{36.5} & - & \textbf{6.7\%} \\ \hline
\end{tabular}
\end{table}

\section{Conclusion and Future Work}
In this work, we present Workspace Guided-DaSH, \blue{a novel hybrid MRMP method for environments with narrow passages.}
Our \blue{method} builds off of a state-of-the-art hybrid planning framework, DaSH, using \blue{workspace} guidance to enable its use for a multi-robot planning problem that lacks a highly structured search space. Additionally, we add support for kinodynamic planning via a modified conflict resolution structure, which also boosts the scalability of our \blue{method} by eliminating unnecessary work when finding motion solutions.

Workspace Guided-DaSH leverages topological guidance in multiple ways, both to move between different levels of composition during planning and to guide sampling through narrow passages in the workspace.
By using knowledge of the workspace, our method is able
to efficiently find paths for robot groups of size up to an order of magnitude larger than existing state-of-the-art methods.
Building off of this work, we plan to explore the use of different kinodynamic low-level motion planners which plan more intelligently for multi-robot systems by interleaving sampling-based planning and trajectory optimization~\cite{moth-ddbcbsfmrkmp-23}.
Additionally, we plan to explore using parallel computing to further improve the performance of this method, taking advantage of the hypergraph's ability to naturally decompose the multi-robot planning problems.



\bibliographystyle{IEEEtran.bst}
\bibliography{robotics.bib}

\end{document}